\documentclass[10pt,a4paper,onecolumn]{article}
\usepackage{marginnote}
\usepackage{graphicx}
\usepackage{xcolor}
\usepackage{authblk,etoolbox}
\usepackage{titlesec}
\usepackage{calc}
\usepackage{tikz}
\usepackage{hyperref}
\hypersetup{colorlinks,breaklinks=true,
            urlcolor=[rgb]{0.0, 0.5, 1.0},
            linkcolor=[rgb]{0.0, 0.5, 1.0}}
\usepackage{caption}
\usepackage{tcolorbox}
\usepackage{amssymb,amsmath}
\usepackage{ifxetex,ifluatex}
\usepackage{seqsplit}
\usepackage{xstring}

\usepackage{float}
\let\origfigure\figure
\let\endorigfigure\endfigure
\renewenvironment{figure}[1][2] {
    \expandafter\origfigure\expandafter[H]
} {
    \endorigfigure
}

\usepackage{fixltx2e} 
\usepackage[
  backend=biber,
]{biblatex}
\bibliography{paper.bib}

\let\textttOrig=\texttt
\def\texttt#1{\expandafter\textttOrig{\seqsplit{#1}}}
\renewcommand{\seqinsert}{\ifmmode
  \allowbreak
  \else\penalty6000\hspace{0pt plus 0.02em}\fi}

\makeatletter
\let\href@Orig=\href
\def\href@Urllike#1#2{\href@Orig{#1}{\begingroup
    \def\Url@String{#2}\Url@FormatString
    \endgroup}}
\def\href@Notdoi#1#2{\def\tempa{#1}\def\tempb{#2}%
  \ifx\tempa\tempb\relax\href@Urllike{#1}{#2}\else
  \href@Orig{#1}{#2}\fi}
\def\href#1#2{%
  \IfBeginWith{#1}{https://doi.org}%
  {\href@Urllike{#1}{#2}}{\href@Notdoi{#1}{#2}}}
\makeatother

\newlength{\cslhangindent}
\newlength{\csllabelwidth}
\newenvironment{CSLReferences}[3] 
 {
  \setlength{\parindent}{0pt}
  \ifodd #1 \everypar{\setlength{\hangindent}{\cslhangindent}}\ignorespaces\fi
  \ifnum #2 > 0
  \setlength{\parskip}{#2\baselineskip}
  \fi
 }%
 {}
\usepackage{calc}

\usepackage[top=3.5cm, bottom=3cm, right=1.5cm, left=1.0cm,
            headheight=2.2cm, reversemp, includemp, marginparwidth=4.5cm]{geometry}

\titleformat{\section}
  {\normalfont\sffamily\Large\bfseries}
  {}{0pt}{}
\titleformat{\subsection}
  {\normalfont\sffamily\large\bfseries}
  {}{0pt}{}
\titleformat{\subsubsection}
  {\normalfont\sffamily\bfseries}
  {}{0pt}{}
\titleformat*{\paragraph}
  {\sffamily\normalsize}

\usepackage{fancyhdr}
\makeatletter
\let\ps@plain\ps@fancy
\fancyheadoffset[L]{4.5cm}
\fancyfootoffset[L]{4.5cm}

\definecolor{linky}{rgb}{0.0, 0.5, 1.0}

\newtcolorbox{repobox}
   {colback=red, colframe=red!75!black,
     boxrule=0.5pt, arc=2pt, left=6pt, right=6pt, top=3pt, bottom=3pt}

\newcommand{\ExternalLink}{%
   \tikz[x=1.2ex, y=1.2ex, baseline=-0.05ex]{%
       \begin{scope}[x=1ex, y=1ex]
           \clip (-0.1,-0.1)
               --++ (-0, 1.2)
               --++ (0.6, 0)
               --++ (0, -0.6)
               --++ (0.6, 0)
               --++ (0, -1);
           \path[draw,
               line width = 0.5,
               rounded corners=0.5]
               (0,0) rectangle (1,1);
       \end{scope}
       \path[draw, line width = 0.5] (0.5, 0.5)
           -- (1, 1);
       \path[draw, line width = 0.5] (0.6, 1)
           -- (1, 1) -- (1, 0.6);
       }
   }

\patchcmd{\@maketitle}{center}{flushleft}{}{}
\patchcmd{\@maketitle}{center}{flushleft}{}{}
\patchcmd{\@maketitle}{\LARGE}{\LARGE\sffamily}{}{}
\def\maketitle{{%
  
  \AB@maketitle}}
\makeatletter
\renewcommand\AB@affilsepx{ \protect\Affilfont}
\renewcommand\AB@affilnote[1]{{\bfseries #1}\hspace{3pt}}
\renewcommand{\affil}[2][]%
   {\newaffiltrue\let\AB@blk@and\AB@pand
      \if\relax#1\relax\def\AB@note{\AB@thenote}\else\def\AB@note{#1}%
        \setcounter{Maxaffil}{0}\fi
        \begingroup
        \let\href=\href@Orig
        \let\texttt=\textttOrig
        \let\protect\@unexpandable@protect
        \def\thanks{\protect\thanks}\def\footnote{\protect\footnote}%
        \@temptokena=\expandafter{\AB@authors}%
        {\def\\{\protect\\\protect\Affilfont}\xdef\AB@temp{#2}}%
         \xdef\AB@authors{\the\@temptokena\AB@las\AB@au@str
         \protect\\[\affilsep]\protect\Affilfont\AB@temp}%
         \gdef\AB@las{}\gdef\AB@au@str{}%
        {\def\\{, \ignorespaces}\xdef\AB@temp{#2}}%
        \@temptokena=\expandafter{\AB@affillist}%
        \xdef\AB@affillist{\the\@temptokena \AB@affilsep
          \AB@affilnote{\AB@note}\protect\Affilfont\AB@temp}%
      \endgroup
       \let\AB@affilsep\AB@affilsepx
}
\makeatother

\renewcommand\Affilfont{\sffamily\small\mdseries}
\ifnum 0\ifxetex 1\fi\ifluatex 1\fi=0 
  \usepackage[T1]{fontenc}
  \usepackage[utf8]{inputenc}

\else 
  \ifxetex
    \usepackage{mathspec}
    \usepackage{fontspec}

  \else
    \usepackage{fontspec}
  \fi
  \defaultfontfeatures{Ligatures=TeX,Scale=MatchLowercase}

\fi
\IfFileExists{upquote.sty}{\usepackage{upquote}}{}
\IfFileExists{microtype.sty}{%
\usepackage{microtype}
\UseMicrotypeSet[protrusion]{basicmath} 
}{}

\usepackage{hyperref}
\hypersetup{unicode=true,
            pdftitle={gym-saturation: an OpenAI Gym environment for saturation provers},
            pdfborder={0 0 0},
            breaklinks=true}
\usepackage{color}
\usepackage{fancyvrb}

\DefineVerbatimEnvironment{Highlighting}{Verbatim}{commandchars=\\\{\}}
\newenvironment{Shaded}{}{}

\newcommand{\BuiltInTok}[1]{#1}

\newcommand{\CommentTok}[1]{\textcolor[rgb]{0.38,0.63,0.69}{\textit{#1}}}

\newcommand{\ControlFlowTok}[1]{\textcolor[rgb]{0.00,0.44,0.13}{\textbf{#1}}}

\newcommand{\DecValTok}[1]{\textcolor[rgb]{0.25,0.63,0.44}{#1}}

\newcommand{\ImportTok}[1]{#1}

\newcommand{\KeywordTok}[1]{\textcolor[rgb]{0.00,0.44,0.13}{\textbf{#1}}}
\newcommand{\NormalTok}[1]{#1}
\newcommand{\OperatorTok}[1]{\textcolor[rgb]{0.40,0.40,0.40}{#1}}

\newcommand{\SpecialCharTok}[1]{\textcolor[rgb]{0.25,0.44,0.63}{#1}}
\newcommand{\SpecialStringTok}[1]{\textcolor[rgb]{0.73,0.40,0.53}{#1}}
\newcommand{\StringTok}[1]{\textcolor[rgb]{0.25,0.44,0.63}{#1}}
\newcommand{\VariableTok}[1]{\textcolor[rgb]{0.10,0.09,0.49}{#1}}

\usepackage{longtable,booktabs}

\let\addcontentslineOrig=\addcontentsline
\def\addcontentsline#1#2#3{\bgroup
  \let\texttt=\textttOrig\addcontentslineOrig{#1}{#2}{#3}\egroup}
\let\markbothOrig\markboth
\def\markboth#1#2{\bgroup
  \let\texttt=\textttOrig\markbothOrig{#1}{#2}\egroup}
\let\markrightOrig\markright
\def\markright#1{\bgroup
  \let\texttt=\textttOrig\markrightOrig{#1}\egroup}

\usepackage{graphicx,grffile}
\makeatletter
\def\maxwidth{\ifdim\Gin@nat@width>\linewidth\linewidth\else\Gin@nat@width\fi}
\def\maxheight{\ifdim\Gin@nat@height>\textheight\textheight\else\Gin@nat@height\fi}
\makeatother
\setkeys{Gin}{width=\maxwidth,height=\maxheight,keepaspectratio}
\IfFileExists{parskip.sty}{%
\usepackage{parskip}
}{
\setlength{\parindent}{0pt}
\setlength{\parskip}{6pt plus 2pt minus 1pt}
}
\providecommand{\tightlist}{%
  \setlength{\itemsep}{0pt}\setlength{\parskip}{0pt}}
\ifx\paragraph\undefined\else
\let\oldparagraph\paragraph
\renewcommand{\paragraph}[1]{\oldparagraph{#1}\mbox{}}
\fi
\ifx\subparagraph\undefined\else
\let\oldsubparagraph\subparagraph
\renewcommand{\subparagraph}[1]{\oldsubparagraph{#1}\mbox{}}
\fi

\title{gym-saturation: an OpenAI Gym environment for saturation provers}

        \author[1]{Boris Shminke}
    
      \affil[1]{Laboratoire J.A. Dieudonné, CNRS and Université Côte
d'Azur, France}
  \date{\vspace{+1ex}}

\begin{document}
\maketitle

\marginpar{

  \begin{flushleft}
  \sffamily\small

  {\bfseries DOI:} \href{https://doi.org/10.21105/joss.03849}{\color{linky}{10.21105/joss.03849}}

  \vspace{2mm}

  {\bfseries Software}
  \begin{itemize}
    \setlength\itemsep{0em}
    \item \href{https://github.com/openjournals/joss-reviews/issues/3849}{\color{linky}{Review}} \ExternalLink
    \item \href{https://github.com/inpefess/gym-saturation}{\color{linky}{Repository}} \ExternalLink
    \item \href{https://doi.org/10.5281/zenodo.6324282}{\color{linky}{Archive}} \ExternalLink
  \end{itemize}

  \vspace{2mm}

  \par\noindent\hrulefill\par

  \vspace{2mm}

  {\bfseries Editor:} \href{https://github.com/VivianePons}{@VivianePons} \ExternalLink \\
  \vspace{1mm}
    {\bfseries Reviewers:}
  \begin{itemize}
  \setlength\itemsep{0em}
    \item \href{https://github.com/lutzhamel}{@lutzhamel}
    \item \href{https://github.com/quickbeam123}{@quickbeam123}
    \end{itemize}
    \vspace{2mm}

  {\bfseries Submitted:} 01 October 2021\\
  {\bfseries Published:} 03 March 2022

  \vspace{2mm}
  {\bfseries License}\\
  Authors of papers retain copyright and release the work under a Creative Commons Attribution 4.0 International License (\href{http://creativecommons.org/licenses/by/4.0/}{\color{linky}{CC BY 4.0}}).

  \end{flushleft}
}

\texttt{gym-saturation} is an OpenAI Gym (Brockman et al., 2016)
environment for reinforcement learning (RL) agents capable of proving
theorems. Currently, only theorems written in a formal language of the
Thousands of Problems for Theorem Provers (TPTP) library (Sutcliffe,
2017) in clausal normal form (CNF) are supported.
\texttt{gym-saturation} implements the `given clause' algorithm (similar
to the one used in Vampire (Kovács \& Voronkov, 2013) and E Prover
(Schulz et al., 2019)). Being written in Python, \texttt{gym-saturation}
was inspired by PyRes (Schulz \& Pease, 2020). In contrast to the
monolithic architecture of a typical Automated Theorem Prover (ATP),
\texttt{gym-saturation} gives different agents opportunities to select
clauses themselves and train from their experience. Combined with a
particular agent, \texttt{gym-saturation} can work as an ATP. Even with
a non trained agent based on heuristics, \texttt{gym-saturation} can
find refutations for 688 (of 8257) CNF problems from TPTP v7.5.0.

\hypertarget{statement-of-need}{%
\section{Statement of need}\label{statement-of-need}}

Current applications of RL to saturation-based ATPs like Enigma (Jakubuv
et al., 2020) or Deepire (Suda, 2021) are similar in that the
environment and the agent are not separate pieces of software but parts
of larger systems that are hard to disentangle. The same is true for non
saturation-based RL-friendly provers too (e.g.~lazyCoP, Rawson \& Reger
(2021)). This monolithic approach hinders free experimentation with
novel machine learning (ML) models and RL algorithms and creates
unnecessary complications for ML and RL experts willing to contribute to
the field. In contrast, for interactive theorem provers, projects like
HOList (Bansal, Loos, Rabe, Szegedy, \& Wilcox, 2019) or GamePad (Huang
et al., 2019) separate the concepts of environment and agent. Such
modular architecture may lead to the development of easily comparable
agents based on diverse approaches (see, e.g. Paliwal et al. (2020) or
Bansal, Loos, Rabe, \& Szegedy (2019)). \texttt{gym-saturation} is an
attempt to implement a modular environment-agent architecture of an
RL-based ATP. In addition, some RL empowered saturation ATPs are not
accompanied with their source code (Abdelaziz et al., 2022), while
\texttt{gym-saturation} is open-source software.

\hypertarget{usage-example}{%
\section{Usage example}\label{usage-example}}

Suppose we want to prove an extremely simple theorem with a very basic
agent. We can do that in the following way:

\begin{Shaded}
\begin{Highlighting}[]
\CommentTok{\# first we create and reset a OpenAI Gym environment}
\ImportTok{from}\NormalTok{ importlib.resources }\ImportTok{import}\NormalTok{ files}
\ImportTok{import}\NormalTok{ gym}

\NormalTok{env }\OperatorTok{=}\NormalTok{ gym.make(}
    \StringTok{"gym\_saturation:saturation{-}v0"}\NormalTok{,}
    \CommentTok{\# we will try to find a proof shorter than 10 steps}
\NormalTok{    step\_limit}\OperatorTok{=}\DecValTok{10}\NormalTok{,}
    \CommentTok{\# for a classical syllogism about Socrates}
\NormalTok{    problem\_list}\OperatorTok{=}\NormalTok{[}
\NormalTok{        files(}\StringTok{"gym\_saturation"}\NormalTok{).joinpath(}
            \StringTok{"resources/TPTP{-}mock/Problems/TST/TST003{-}1.p"}
\NormalTok{        )}
\NormalTok{    ],}
\NormalTok{)}
\NormalTok{env.reset()}
\CommentTok{\# we can render the environment (that will become the beginning of the proof)}
\BuiltInTok{print}\NormalTok{(}\StringTok{"starting hypotheses:"}\NormalTok{)}
\BuiltInTok{print}\NormalTok{(env.render(}\StringTok{"human"}\NormalTok{))}
\CommentTok{\# our \textquotesingle{}age\textquotesingle{} agent will always select clauses for inference}
\CommentTok{\# in the order they appeared in current proof attempt}
\NormalTok{action }\OperatorTok{=} \DecValTok{0}
\NormalTok{done }\OperatorTok{=} \VariableTok{False}
\ControlFlowTok{while} \KeywordTok{not}\NormalTok{ done:}
\NormalTok{    observation, reward, done, info }\OperatorTok{=}\NormalTok{ env.step(action)}
\NormalTok{    action }\OperatorTok{+=} \DecValTok{1}
\CommentTok{\# SaturationEnv has an additional method}
\CommentTok{\# for extracting only clauses which became parts of the proof}
\CommentTok{\# (some steps were unnecessary to find the proof)}
\BuiltInTok{print}\NormalTok{(}\StringTok{"refutation proof:"}\NormalTok{)}
\BuiltInTok{print}\NormalTok{(env.tstp\_proof)}
\BuiltInTok{print}\NormalTok{(}\SpecialStringTok{f"number of attempted steps: }\SpecialCharTok{\{}\NormalTok{action}\SpecialCharTok{\}}\SpecialStringTok{"}\NormalTok{)}
\end{Highlighting}
\end{Shaded}

The output of this script includes a refutation proof found:

\begin{verbatim}
starting hypotheses:
cnf(p_imp_q, hypothesis, ~man(X0) | mortal(X0)).
cnf(p, hypothesis, man(socrates)).
cnf(q, hypothesis, ~mortal(socrates)).
refutation proof:
cnf(_0, hypothesis, mortal(socrates), inference(resolution, [], [p_imp_q, p])).
cnf(_2, hypothesis, $false, inference(resolution, [], [q, _0])).
number of attempted steps: 6
\end{verbatim}

\hypertarget{architecture}{%
\section{Architecture}\label{architecture}}

\texttt{gym-saturation} includes several sub-packages:

\begin{itemize}
\tightlist
\item
  parsing (happens during \texttt{env.reset()} in example code snippet)
\item
  logic operations (happen during \texttt{env.step(action)} in the
  example)
\item
  AI Gym environment implementation
\item
  agent testing (a bit more elaborated version of the \texttt{while}
  loop from the examle)
\end{itemize}

\texttt{gym-saturation} relies on a deduction system of four rules which
is known to be refutationally complete (Brand, 1975):

\begin{align*}
{\frac{C_1\vee A_1,C_2\vee\neg A_2}{\sigma\left(C_1\vee C_2\right)}},\sigma=mgu\left(A_1,A_2\right)\quad\text{(resolution)}
\end{align*} \begin{align*}
{\frac{C_1\vee s\approx t,C_2\vee L\left[r\right]}{\sigma\left(L\left[t\right]\vee C_1\vee C_2\right)}},\sigma=mgu\left(s,r\right)\quad\text{(paramodulation)}
\end{align*} \begin{align*}
{\frac{C\vee A_1\vee A_2}{\sigma\left(C\vee L_1\right)}},\sigma=mgu\left(A_1,A_2\right)\quad\text{(factoring)}
\end{align*} \begin{align*}
\frac{C\vee s\not\approx t}{\sigma\left(C\right)},\sigma=mgu\left(s,t\right)\quad\text{(reflexivity resolution)}
\end{align*}

where \(C,C_1,C_2\) are clauses, \(A_1,A_2\) are atomic formulae, \(L\)
is a literal, \(r,s,t\) are terms, and \(\sigma\) is a substitution
(most general unifier). \(L\left[t\right]\) is a result of substituting
the term \(t\) in \(L\left[r\right]\) for the term \(r\) at only one
chosen position.

For parsing, we use the LARK parser (Shinan, 2021). We represent the
clauses as Python classes forming tree-like structures.
\texttt{gym-saturation} also includes a JSON serializer/deserializer for
those trees. For example, a TPTP clause

\begin{verbatim}
cnf(a2,hypothesis,
    ( ~ q(a) | f(X) = X )).
\end{verbatim}

becomes

\begin{Shaded}
\begin{Highlighting}[]
\NormalTok{Clause(}
\NormalTok{    literals}\OperatorTok{=}\NormalTok{[}
\NormalTok{        Literal(}
\NormalTok{            negated}\OperatorTok{=}\VariableTok{True}\NormalTok{,}
\NormalTok{            atom}\OperatorTok{=}\NormalTok{Predicate(}
\NormalTok{                name}\OperatorTok{=}\StringTok{"q"}\NormalTok{, arguments}\OperatorTok{=}\NormalTok{[Function(name}\OperatorTok{=}\StringTok{"a"}\NormalTok{, arguments}\OperatorTok{=}\NormalTok{[])]}
\NormalTok{            ),}
\NormalTok{        ),}
\NormalTok{        Literal(}
\NormalTok{            negated}\OperatorTok{=}\VariableTok{False}\NormalTok{,}
\NormalTok{            atom}\OperatorTok{=}\NormalTok{Predicate(}
\NormalTok{                name}\OperatorTok{=}\StringTok{"="}\NormalTok{,}
\NormalTok{                arguments}\OperatorTok{=}\NormalTok{[}
\NormalTok{                    Function(name}\OperatorTok{=}\StringTok{"f"}\NormalTok{, arguments}\OperatorTok{=}\NormalTok{[Variable(name}\OperatorTok{=}\StringTok{"X"}\NormalTok{)]),}
\NormalTok{                    Variable(name}\OperatorTok{=}\StringTok{"X"}\NormalTok{),}
\NormalTok{                ],}
\NormalTok{            ),}
\NormalTok{        ),}
\NormalTok{    ],}
\NormalTok{    label}\OperatorTok{=}\StringTok{"a2"}\NormalTok{,}
\NormalTok{)}
\end{Highlighting}
\end{Shaded}

This grammar serves as the glue for \texttt{gym-saturation}
sub-packages, which are, in principle, independent of each other. After
switching to another parser or another deduction system, the agent
testing script won't break, and RL developers won't need to modify their
agents for compatibility (for them, the environment will have the same
standard OpenAI Gym API).

\begin{figure}
\centering
\includegraphics{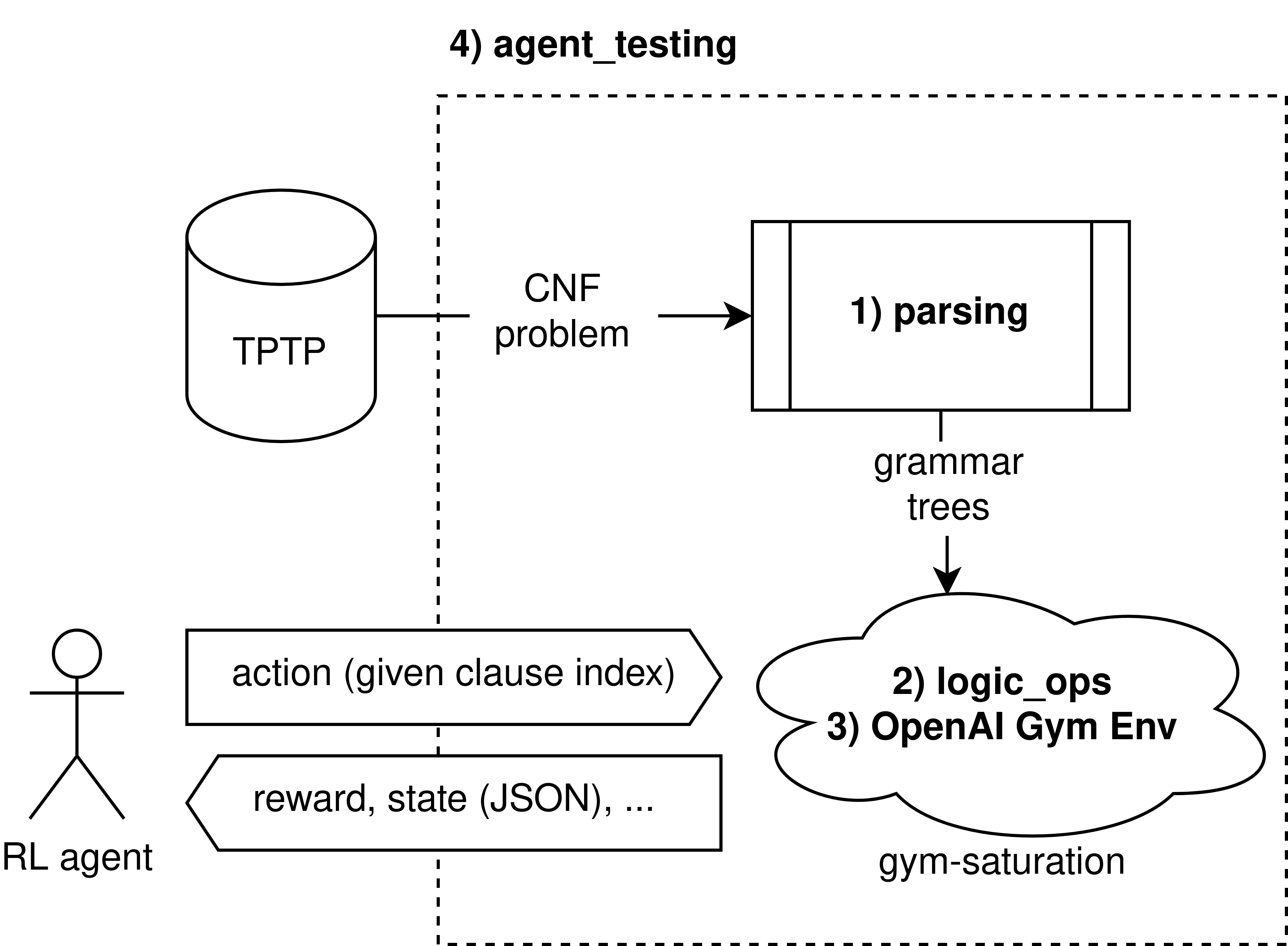}
\caption{A diagram showing interactions between four main subpackages of
\texttt{gym-saturation}: 1) parsing; 2) logic operations (including the
given clause algorithm); 3) OpenAI Gym Env implementation; 4) the agent
testing script.\label{fig:architecture}}
\end{figure}

Agent testing is a simple episode pipeline (see
\autoref{fig:architecture}). It is supposed to be run in parallel
(e.g.~using GNU Parallel, Tange (2021)) for a testing subset of
problems. See the following table for the testing results of two popular
heuristic-based agents on TPTP v7.5.0 (trained RL agents should strive
to be more successful than those primitive baselines):

\begin{longtable}[]{@{}llll@{}}
\toprule
& \textbf{size agent} & \textbf{age agent} & \textbf{size\&age
agent}\tabularnewline
\midrule
\endhead
\textbf{proof found} & 509 & 206 & 688\tabularnewline
\textbf{step limit} & 1385 & 35 & 223\tabularnewline
\textbf{out of memory} & 148 & 149 & 148\tabularnewline
\textbf{5 min time out} & 6215 & 7867 & 7198\tabularnewline
\textbf{total} & 8257 & 8257 & 8257\tabularnewline
\bottomrule
\end{longtable}

\texttt{size\ agent} is an agent which always selects the shortest
clause.

\texttt{age\ agent} is an agent which always selects the clause which
arrived first to the set of unprocessed clauses (`the oldest one').

\texttt{size\&age\ agent} is an agent which selects the shortest clause
five times in a row and then one time --- the oldest one.

`Step limit' means an agent didn't find proof after 1000 steps (the
longest proof found consists of 287 steps). This can work as a `soft
timeout.'

\hypertarget{mentions}{%
\section{Mentions}\label{mentions}}

At the moment of writing this paper, \texttt{gym-saturation} was used by
its author during their PhD studies for creating experimental RL-based
ATPs.

\hypertarget{acknowledgements}{%
\section{Acknowledgements}\label{acknowledgements}}

This work has been supported by the French government, through the 3IA
Côte d'Azur Investments in the Future project managed by the National
Research Agency (ANR) with the reference number ANR-19-P3IA-0002. This
work was performed using HPC resources from GENCI-IDRIS (Grant
2021-AD011013125).

\hypertarget{references}{%
\section*{References}\label{references}}
\addcontentsline{toc}{section}{References}

\hypertarget{refs}{}
\begin{CSLReferences}{1}{0}
\leavevmode\hypertarget{ref-9669114}{}%
Abdelaziz, I., Crouse, M., Makni, B., Austel, V., Cornelio, C., Ikbal,
S., Kapanipathi, P., Makondo, N., Srinivas, K., Witbrock, M., \& Fokoue,
A. (2022). Learning to guide a saturation-based theorem prover.
\emph{IEEE Transactions on Pattern Analysis and Machine Intelligence},
1--1. \url{https://doi.org/10.1109/TPAMI.2022.3140382}

\leavevmode\hypertarget{ref-DBLP:journalsux2fcorrux2fabs-1905-10501}{}%
Bansal, K., Loos, S. M., Rabe, M. N., \& Szegedy, C. (2019). Learning to
reason in large theories without imitation. \emph{CoRR},
\emph{abs/1905.10501}. \url{http://arxiv.org/abs/1905.10501}

\leavevmode\hypertarget{ref-DBLP:confux2ficmlux2fBansalLRSW19}{}%
Bansal, K., Loos, S. M., Rabe, M. N., Szegedy, C., \& Wilcox, S. (2019).
HOList: An environment for machine learning of higher order logic
theorem proving. In K. Chaudhuri \& R. Salakhutdinov (Eds.),
\emph{Proceedings of the 36th international conference on machine
learning, {ICML} 2019, 9-15 june 2019, long beach, california, {USA}}
(Vol. 97, pp. 454--463). {PMLR}.
\url{http://proceedings.mlr.press/v97/bansal19a.html}

\leavevmode\hypertarget{ref-doi:10.1137ux2f0204036}{}%
Brand, D. (1975). Proving theorems with the modification method.
\emph{SIAM Journal on Computing}, \emph{4}(4), 412--430.
\url{https://doi.org/10.1137/0204036}

\leavevmode\hypertarget{ref-DBLP:journalsux2fcorrux2fBrockmanCPSSTZ16}{}%
Brockman, G., Cheung, V., Pettersson, L., Schneider, J., Schulman, J.,
Tang, J., \& Zaremba, W. (2016). OpenAI gym. \emph{CoRR},
\emph{abs/1606.01540}. \url{http://arxiv.org/abs/1606.01540}

\leavevmode\hypertarget{ref-DBLP:confux2ficlrux2fHuangDSS19}{}%
Huang, D., Dhariwal, P., Song, D., \& Sutskever, I. (2019). GamePad: {A}
learning environment for theorem proving. \emph{7th International
Conference on Learning Representations, {ICLR} 2019, New Orleans, LA,
USA, May 6-9, 2019}. \url{https://openreview.net/forum?id=r1xwKoR9Y7}

\leavevmode\hypertarget{ref-DBLP:confux2fcadeux2fJakubuvCOP0U20}{}%
Jakubuv, J., Chvalovský, K., Olsák, M., Piotrowski, B., Suda, M., \&
Urban, J. (2020). {ENIGMA} anonymous: Symbol-independent inference
guiding machine (system description). In N. Peltier \& V.
Sofronie-Stokkermans (Eds.), \emph{Automated reasoning - 10th
international joint conference, {IJCAR} 2020, paris, france, july 1-4,
2020, proceedings, part {II}} (Vol. 12167, pp. 448--463). Springer.
\url{https://doi.org/10.1007/978-3-030-51054-1_29}

\leavevmode\hypertarget{ref-DBLP:confux2fcavux2fKovacsV13}{}%
Kovács, L., \& Voronkov, A. (2013). First-order theorem proving and
vampire. In N. Sharygina \& H. Veith (Eds.), \emph{Computer aided
verification - 25th international conference, {CAV} 2013, saint
petersburg, russia, july 13-19, 2013. proceedings} (Vol. 8044, pp.
1--35). Springer. \url{https://doi.org/10.1007/978-3-642-39799-8_1}

\leavevmode\hypertarget{ref-DBLP:confux2faaaiux2fPaliwalLRBS20}{}%
Paliwal, A., Loos, S. M., Rabe, M. N., Bansal, K., \& Szegedy, C.
(2020). Graph representations for higher-order logic and theorem
proving. \emph{The Thirty-Fourth {AAAI} Conference on Artificial
Intelligence, {AAAI} 2020, the Thirty-Second Innovative Applications of
Artificial Intelligence Conference, {IAAI} 2020, the Tenth {AAAI}
Symposium on Educational Advances in Artificial Intelligence, {EAAI}
2020, New York, NY, USA, February 7-12, 2020}, 2967--2974.
\url{https://aaai.org/ojs/index.php/AAAI/article/view/5689}

\leavevmode\hypertarget{ref-DBLP:confux2ftableauxux2fRawsonR21}{}%
Rawson, M., \& Reger, G. (2021). lazyCoP: Lazy paramodulation meets
neurally guided search. In A. Das \& S. Negri (Eds.), \emph{Automated
reasoning with analytic tableaux and related methods - 30th
international conference, {TABLEAUX} 2021, birmingham, UK, september
6-9, 2021, proceedings} (Vol. 12842, pp. 187--199). Springer.
\url{https://doi.org/10.1007/978-3-030-86059-2_11}

\leavevmode\hypertarget{ref-DBLP:confux2fcadeux2f0001CV19}{}%
Schulz, S., Cruanes, S., \& Vukmirovic, P. (2019). Faster, higher,
stronger: {E} 2.3. In P. Fontaine (Ed.), \emph{Automated deduction -
{CADE} 27 - 27th international conference on automated deduction, natal,
brazil, august 27-30, 2019, proceedings} (Vol. 11716, pp. 495--507).
Springer. \url{https://doi.org/10.1007/978-3-030-29436-6_29}

\leavevmode\hypertarget{ref-DBLP:confux2fcadeux2f0001P20}{}%
Schulz, S., \& Pease, A. (2020). Teaching automated theorem proving by
example: PyRes 1.2 - (system description). In N. Peltier \& V.
Sofronie-Stokkermans (Eds.), \emph{Automated reasoning - 10th
international joint conference, {IJCAR} 2020, paris, france, july 1-4,
2020, proceedings, part {II}} (Vol. 12167, pp. 158--166). Springer.
\url{https://doi.org/10.1007/978-3-030-51054-1_9}

\leavevmode\hypertarget{ref-LARK}{}%
Shinan, E. (2021). \emph{Lark-parser} (Version 0.12.0).
\url{https://pypi.org/project/lark-parser/}

\leavevmode\hypertarget{ref-DBLP:confux2fcadeux2f000121a}{}%
Suda, M. (2021). Improving ENIGMA-style clause selection while learning
from history. In A. Platzer \& G. Sutcliffe (Eds.), \emph{Automated
deduction - {CADE} 28 - 28th international conference on automated
deduction, virtual event, july 12-15, 2021, proceedings} (Vol. 12699,
pp. 543--561). Springer.
\url{https://doi.org/10.1007/978-3-030-79876-5_31}

\leavevmode\hypertarget{ref-Sut17}{}%
Sutcliffe, G. (2017). {The TPTP Problem Library and Associated
Infrastructure. From CNF to TH0, TPTP v6.4.0}. \emph{Journal of
Automated Reasoning}, \emph{59}(4), 483--502.

\leavevmode\hypertarget{ref-tange_2021_5233953}{}%
Tange, O. (2021). \emph{GNU parallel 20210822 ('kabul')}. Zenodo.
\url{https://doi.org/10.5281/zenodo.5233953}

\end{CSLReferences}

\end{document}